\documentclass[runningheads]{llncs}
\usepackage{graphicx}
\usepackage{amssymb}
\usepackage{algorithm}
\usepackage{balance}
\usepackage{algorithmic}
\usepackage{cite}

\usepackage{amsmath}
\usepackage{amssymb}
\usepackage{subfigure}

\usepackage{float}
\usepackage{wrapfig}

\usepackage{color}
\usepackage{booktabs}
\usepackage{threeparttable}
\usepackage{multirow}
\usepackage{makecell}

\begin{document}
\title{Long Short-Term Attention}
\titlerunning{Long Short-Term Attention}
%
\author{Guoqiang Zhong\inst{1} \and
 Xin Lin \inst{1} \and
Kang Chen \inst{1}\and
Qingyang Li  \inst{1}\and
Kaizhu Huang \inst{2}}
\authorrunning{G. Zhong et al.}
%
\institute{
 Department of Computer Science and Technology, Ocean University of China, Qingdao 266100, China \\
\email{gqzhong@ouc.edu.cn; 2410767409@qq.com; chenkoucer@qq.com; 1194094543@qq.com;}\\
\and
 Department of Electrical and Electronic Engineering, Xi¡¯an Jiaotong-Liverpool University, SIP, Suzhou 215123, China \\
 \email{Kaizhu.Huang@xjtlu.edu.cn}
 }

\maketitle
%
\begin{abstract}
Attention is an important cognition process of humans, which helps humans concentrate on critical information during their perception and learning. However, although many machine learning models can remember information of data, they have no the attention mechanism. For example, the long short-term memory (LSTM) network is able to remember sequential information, but it cannot pay special attention to part of the sequences. In this paper, we present a novel model called long short-term attention (LSTA), which seamlessly integrates the attention mechanism into the inner cell of LSTM. More than processing long short term dependencies, LSTA can focus on important information of the sequences with the attention mechanism. Extensive experiments demonstrate that LSTA outperforms LSTM and related models on the sequence learning tasks.
\keywords{Machine learning \and Sequence learning \and Attention mechanism \and Long short-term memory \and Long short-term attention.}
\end{abstract}
\section{Introduction}

With the attention mechanism, human can naturally focus on vital information and ignore irrelevant information during one's perception and cognition~\cite{cognitive,show}. Based on this fact, many brain-inspired learning models have been deeply studied and widely applied in recent years~\cite{Brain-Inspired,CognitiveComputation,Modelling,BiologicallyInspired,VisualAttention,Where,bottom,videosatt}. However, although many machine learning models can learn effective representations of data and memorize the data information, they cannot pay attention to important part of the data. For instance, long short-term memory (LSTM) \cite{lstmone} is a widely used model for sequence learning. However, it lacks the attention mechanism. To address this problem, some work tries to apply the attention mechanism to LSTM. Nevertheless, most of these models only add the attention mechanisms outside the LSTM cells and have not thoroughly solved the issue that LSTM have no the attention mechanism itself \cite{show,effective,self-attentive}.

In this paper, we propose a novel model called long short term attention (LSTA), which seamlessly integrates the attention mechanism into the inner cell of LSTM. In this case, LSTA can simultaneously remember historical information and notice crucial details in the sequences. In the experiments for sequence learning, we demonstrate the advantage of LSTA over LSTM.

The rest of this paper is organized as follows. In Section~\ref{Related Work}, we introduce some previous work related to LSTA, including LSTM and some models using the attention mechanism. In Section~\ref{Architecture}, we present LSTA in detail. In Section~\ref{Experiments}, we report the experimental results on two sequence learning tasks, i.e. image classification and sentiment analysis. Section~\ref{Conclusion} concludes this paper.

\section{Related Work}
\label{Related Work}
In this section, we review some previous work related to LSTA, including LSTM and several models using the attention mechanism.

\subsection{LSTM}
\label{Normative}

LSTM is a powerful learning model for sequential data and has been widely applied in many areas, such as speech recognition and handwritten character recognition~\cite{LSTM, Bidirectional}. The cell of LSTM includes an input gate, a forget gate and an output gate. These gate and the state of the cell can be updated as follows:
\begin{equation}\label{equation1}
 \boldsymbol{{f}_{t}=\sigma({W}_{f}[h_{t-1},x_t]+{b}_{f})},
 \end{equation}
 \begin{equation}
 \boldsymbol{{i}_{t}=\sigma({W}_{i}[h_{t-1},x_t]+{b}_{i})},\\
 \end{equation}
 \begin{equation}
   \boldsymbol{\tilde{C}_{t}=\tanh({W}_{\tilde{c}}[h_{t-1},x_t]+{b}_{\tilde{c}})},\\
   \end{equation}
   \begin{equation}
   \boldsymbol{{C}_{t}={f}_{t}\ast {C}_{t-1}+{i}_{t}\ast\tilde{C}_{t}},\\
   \end{equation}
   \begin{equation}
   \boldsymbol{{o}_{t}=\sigma({W}_{o}[h_{t-1},x_t]+{b}_{o})},\\
   \end{equation}
   \begin{equation}
   \boldsymbol{{h}_{t}={o}_{t}\ast\tanh({C}_{t})}.\\
 \end{equation}
Here, $\boldsymbol{{W}_{f},{W}_{i},{W}_{\tilde{c}},{W}_{o}}$ are the weight parameters and $\boldsymbol{{b}_{f},{b}_{i},{b}_{\tilde{c}},{b}_{o}}$ are biases.
The forget gate $\boldsymbol{{f}_{t}}$ primarily controls the cell state by forgetting the previous moment information. In a similar way, the input gate $\boldsymbol{{i}_{t}}$ and output gate $\boldsymbol{{o}_{t}}$ control the information that will be input to the LSTM cell and output at the current moment, respectively. These three gates are crucial parts of the LSTM cell, which is used to update the current state of the LSTM cell $\boldsymbol{{C}_{t}}$ and obtain new cell output $\boldsymbol{{h}_{t}}$. 

In order to optimize the performance of LSTM, many extensions of LSTM have been proposed recently~\cite{speech, wider, Sequentially, phased}. In \cite{predrnn}, the spatiotemporal LSTM (ST-LSTM) units are designed for memorizing both spatial and temporal information. \cite{convolutional} introduces a convolutional LSTM (ConvLSTM), which extends the fully connected LSTM to have convolutional architectures in both the input-to-gate and gate-to-gate transitions. In addition, \cite{wider} introduces a tensorized LSTM model, which represent the hidden states with tensors.

As discussed above, LSTM and most of its extensions mainly focus on processing the sequential data, but cannot pay attention to the important information in the sequences. In this paper, we present a model that can integrate the attention mechanism into the inner-cell of LSTM.

\subsection{Models Using the Attention Mechanism}

The primary function of the attention mechanism is selection and allocation \cite{cognitive, Where}. It leads to quick processing of information, with an efficient information choice and concentration of the computing power on the crucial tasks \cite{cognitive}. \cite{control} introduces the attention mechanism in the human cognitive system, with which human pays attention to the noteworthy information and ignores irrespective information \cite{show,control,AttentionBased}. In the cognitive computation area, the attention mechanism has been widely applied, such as the work to resolve the human visual neural computational problem~\cite{CognitiveAttention} and that to model the retrieval mechanism of associations from the associative memory \cite{TheRole}.

In particular, a large amount of attention based deep learning models have been proposed in recent years. For example, \cite{structured} presents the structured attention networks, which incorporate graphical models to generalize simple attention. Alternatively, \cite{attention} introduces a self-attention mechanism model, which is applied to replace the common recurrent and convolutional models. It relies entirely on the attention mechanism to compute representations of its input and output. Moreover, in \cite{Modelling}, the selective attention for identification model (SAIM) is applied to visual search applications. The SAIM simulates the human ability to complete translation invariant recognition of multiple scenes. Additionally, in \cite{end}, a recurrent attention mechanism network is proposed. It is an end-to-end memory learning model used on several language modeling tasks.

As mentioned above, many attention based methods have been proposed to address visual or language processing problems. However, rare work has integrated the attention mechanism into the cell of LSTM to improve its performance in sequence learning.

\section{Long Short-Term Attention}
\label{Architecture}
In this section, we introduce the proposed long short-term attention (LSTA) model in detail, which seamlessly integrates the attention mechanism into the cell of LSTM. For clarity, we first introduce the added attention gate in Sec. \ref{AttentionBlock}, and then the architecture and learning of LSTA in Sec. \ref{LSTA}.

\begin{figure}[h]
  \centering
  \includegraphics[width=3in]{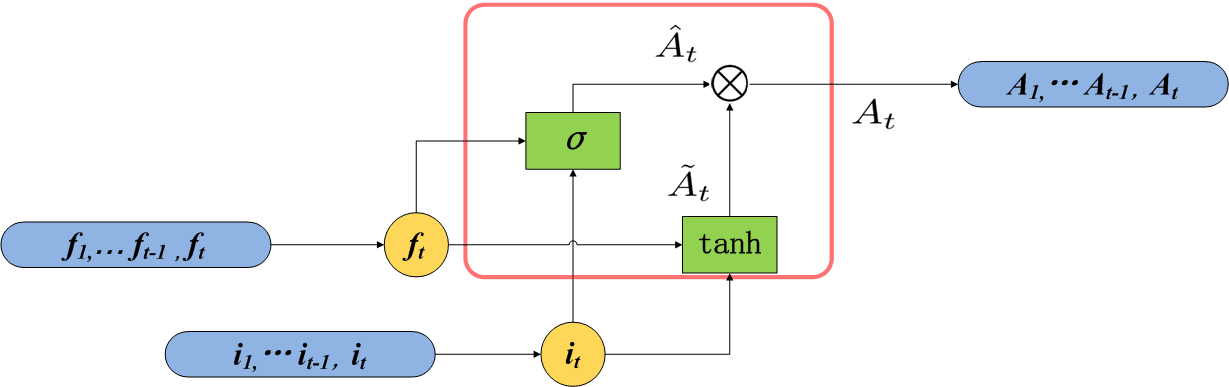}
  \caption{An illustration of the attention gate.}
  \label{Attentionblock}
\end{figure}

\subsection{The Attention Gate}\label{AttentionBlock}

Fig. \ref{Attentionblock} shows the structure of the attention gate of LSTA, which accepts the inputs from the input gate and the forget gate. Eq. (\ref{equation4}) is the update formula of the attention gate:
\begin{equation}\label{equation4}
\boldsymbol{{A}_{t}}=\boldsymbol{\psi(\hat{A}_{t}[f_t, i_t], \tilde{A}_{t}[f_t, i_t])}=\boldsymbol{\hat{A}_{t}\bigotimes\tilde{A}_{t}},
\end{equation}
where $\boldsymbol{\hat{A}_{t}}$ and $\boldsymbol{\tilde{A}_{t}}$ are defined as follows:
\begin{equation}\label{equation23}
  \boldsymbol{\hat{A}_{t}=\sigma({W}_{\hat{a}}[f_t, i_t]+{b}_{\hat{a}})},\\
  \end{equation}
  \begin{equation}
   \boldsymbol{\tilde{A}_{t}=\tanh({W}_{\tilde{a}}[f_t, i_t]+{b}_{\tilde{a}})}.
\end{equation}

Here, $\boldsymbol{{W}_{\tilde{a}}}$ and $\boldsymbol{{W}_{\hat{a}}}$ are weight parameters, while $\boldsymbol{{b}_{\tilde{a}}}$ and $\boldsymbol{{b}_{\hat{a}}}$ are biases. The sigmoid function $\boldsymbol{\sigma}$ is employed to compute $\boldsymbol{\hat{A}_{t}}$, which indicates the ratios of the attention elements as shown in Eq. (\ref{equation4}). Similarly, the $\boldsymbol{\tanh}$ function is used to get the candidate attention values $\boldsymbol{\tilde{A}_{t}}$, which can be positive or negative.

In Eq. (\ref{equation4}), $\bigotimes$ represents the element-wise multiplication. We multiply the elements between $\boldsymbol{\tilde{A}_{t}}$  and $\boldsymbol{\hat{A}_{t}}$ to obtain the output of the attention gate $\boldsymbol{{A}_{t}}$. The attention gate determines the attention distribution on the information at the current cell. In the following, we introduce how it can be seamlessly integrated into the cell of LSTM.

\subsection{LSTA}\label{LSTA}

\begin{figure}[h]
 \centering
\includegraphics[width=2.5in]{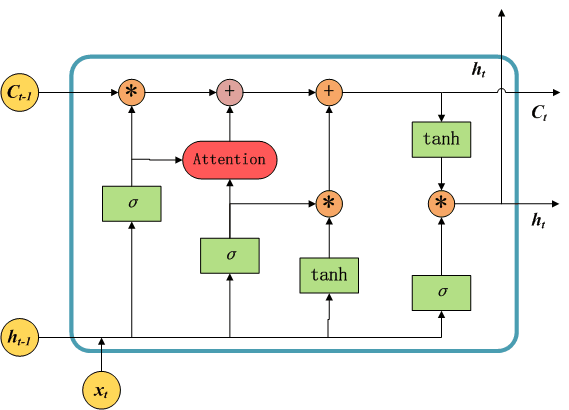}
\caption{The LSTA cell. The module with red color is the attention gate.}
\label{figure2}
\end{figure}

In order to endow the attention mechanism to LSTM, we propose the LSTA model which integrates the attention gate introduced above inside the LSTM cell. Fig. \ref{figure2} is a diagram of the LSTA cell. Particularly, LSTA can pay attention to important information in the sequences during its learning process.

LSTA inherits the three gates of LSTM. For the update of its cell state, we can compute it as
\begin{equation}\label{equation5}
 \begin{split}
   \boldsymbol{\hat{C}_{t}}&=\boldsymbol{{C}_{t}+{A}_{t}}\\
               &=\boldsymbol{{f}_{t}\ast{C}_{t-1}+{i}_{t}\ast\tilde{C}_{t}+{A}_{t}}\\
               &=\boldsymbol{{f}_{t}\ast{C}_{t-1}+{i}_{t}\ast\tilde{C}_{t}+\boldsymbol{\tilde{A}_{t}\bigotimes\hat{A}_{t}}}.\\
\end{split}
\end{equation}
Here, $\boldsymbol{{A}_{t}}$ is the output of the attention gate and $\boldsymbol{{C}_{t}}$ is the original LSTM cell state. In this case, we integrate the attention mechanism into LSTM unit, such that the new model, LSTA, can not only memorize the sequential information, but also pay attention to important information in the sequences.

Accordingly, the output gate of LSTA can be updated as
\begin{align}
\label{equationh}
  &\boldsymbol{{h}_{t}={o}_{t}\ast\tanh(\hat{C}_{t})}.
\end{align}

Note that, in LSTA, the attention mechanise is applied inside the LSTM cell, unlike previous attention based LSTM models, in which attention is added after the whole sequence has been handled by all the LSTM cells. Therefore, LSTA is quite different from LSTM and most of its attention based variants. LSTA enables the sequence learning to focus on important parts of the input data and automatically ignore irrelevant parts, so as to improve its performance.


\section{Experiments}\label{Experiments}

To evaluate the proposed model LSTA, we have conducted extensive experiments on two sequence learning tasks, image classification and semantic analysis. In the following, we report the experimental settings and results.

\subsection{Experiments on the Image Classification Task}
\label{Image}
In this section, we used the MNIST and Fashion-MNIST data sets to test the performance of LSTA. MNIST is a handwritten digit data set. It contains seventy thousands of $28 \times 28$ gray scale images, which belong to 10 classes. For all the images, 60,000 are used for training and the other 10,000 for test \cite{mnist}. Alternatively, Fashion-MNIST is an image data set, while its image format and number are both the same as the MNIST data set \cite{fashion}. In our experiments, we considered the rows of an image as sequential data to perform image classification.

For testing the performance of LSTA, we compared it with some relevant models. As LSTA integrates the attention mechanism into the LSTM cell, the most closely related model to LSTA is LSTM. Hence, we set LSTM as our baseline. Furthermore, we also compared LSTA with gated recurrent unit (GRU)~\cite{ChungGCB14}, bidirectional LSTM (Bi-LSTM)~\cite{graves2005framewise} and nested LSTM (NLSTM)~\cite{NLSTM} in our experiments. Note that, although there are many attention based variants of LSTM, they are quite different from LSTA. We can also apply the attention mechanism outside the LSTA cell as same as them. Hence, we have not compared with them in our work.

Table \ref{tableFS} shows the image classification results obtained by LSTA and the compared models on the MNIST and Fashion-MNIST data sets. As we can see, LSTA outperforms all the compared models consistently. This demonstrate the advantage of LSTA over LSTM and its variants and the importance of the attention mechanism during sequence learning. \vspace{-0.3cm}

\begin{table*}[h]
  \caption{Accuracy obtained by LSTA and related models on the MNIST and Fashion-MNIST data sets.}
  \smallskip
  \centering
   \setlength{\tabcolsep}{0.85mm}{
  \begin{tabular}{l|ccccc}
    \toprule
     Data set  &LSTM &GRU &Bi-LSTM &NLSTM&\textbf{LSTA}  \\
 \midrule
   MNIST  &97.47\% &97.79\% &97.81\% &97.75\% & $\boldsymbol{97.85\%}$ \\
   \midrule
   Fashion-MNIST &87.46\% &88.16\% &88.18\% &88.32\% & $\boldsymbol{88.60\%}$  \\
    \bottomrule
  \end{tabular}
}\label{tableFS}
\end{table*}\vspace{-0.3cm}

\begin{figure}[h]
 \centering
\subfigure[Accuracy curves] { \label{fig:a1}
\includegraphics[width=0.45\columnwidth]{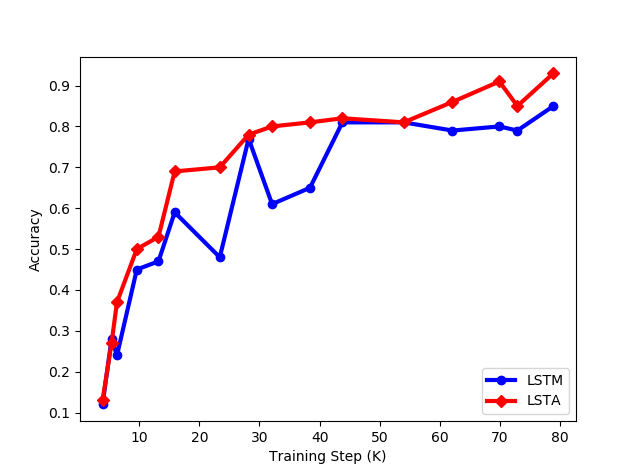}
}
\subfigure[Loss curves] { \label{fig:b2}
\includegraphics[width=0.45\columnwidth]{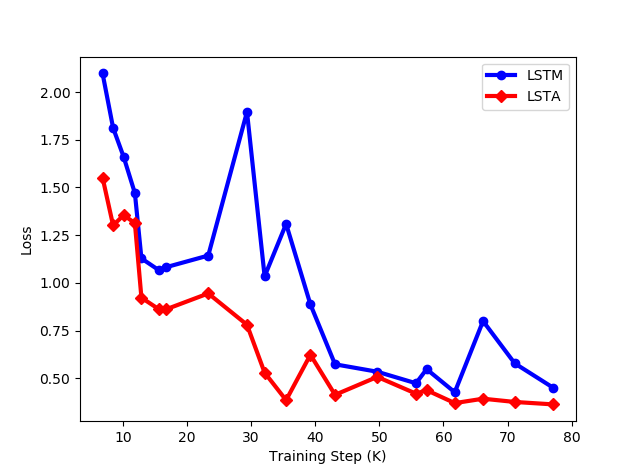}
}
\caption{The accuracy and loss curves of LSTM and LSTA on the Fashion-MNIST data set. (a) shows the accuracy curves; and (b) shows the loss curves.}\label{QualitativeF}
\end{figure}

To further analyze the advantage of LSTA over LSTM, we draw the learning curves of LSTM and LSTA obtained on the Fashion-MNIST data set in Fig. \ref{QualitativeF}. Note that, we used the same loss function for LSTM and LSTA in our experiments. Fig. \ref{QualitativeF} (a) shows the accuracy curves against the training steps, while Fig. \ref{QualitativeF} (b) shows the loss curves against the training steps. As can be seen, due to the attention mechanism, LSTA consistently performs better, and converges faster than LSTM.\vspace{-0.7cm}

\vspace{-0.6cm}

\subsection{Experiments on the Sentiment Analysis Task}
\label{sentimentanalysis}

Sentiment analysis is an interesting and important learning task \cite{SentimentAnalysis,SemEval14,Twitter}. In order to evaluate the performance of LSTA, we conducted experiments on both the classical sentiment analysis and aspect based sentiment analysis.

\subsubsection{Classical Sentiment Analysis}
\label{IMDB}

In this experiment, we used the internet movie review database (IMDB) \cite{SentimentAnalysis} to test LSTA on classical sentiment analysis. IMDB is a crawler data set about the internet movie reviews. Based on the emotion of the reviews, it divides all the film reviews into the positive and negative categories.

\begin{figure}[h]
 \centering
\subfigure[Error rate] { \label{figureimdb}
\includegraphics[width=0.45\columnwidth]{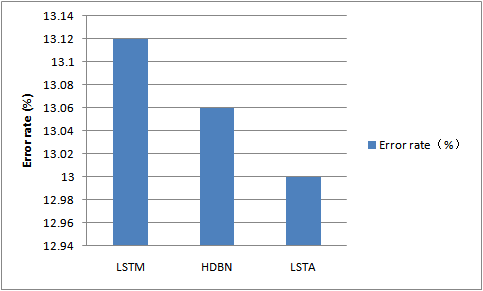}
}
\subfigure[Running time] { \label{figureruntime}
\includegraphics[width=0.45\columnwidth]{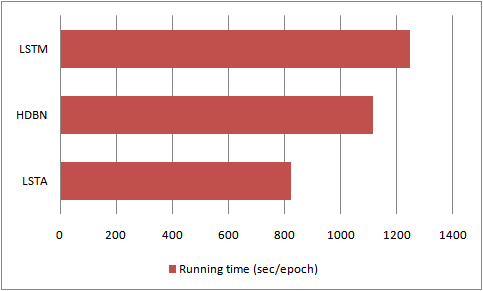}
}\vspace{-0.2cm}
\caption{The error rate and the running time of LSTM, HDBN and LSTA on the IMDB data set. (a) shows the error rate obtained by the three models; (b) depicts their running time.}
\label{QualitativeF2}
\end{figure}

In our work, we compared LSTA with LSTM and hybrid deep belief network (HDBN) \cite{yan2015learning}, which is an effective deep network for sentiment analysis. The error rate and the running time of LSTA and the compared models are depicted in Fig. \ref{figureimdb} and \ref{figureruntime}, respectively.
As we can see, among the compared model, LSTA obtained the best classification accuracy and used the least running time.\vspace{-0.5cm}
\vspace{-0.5cm}

\subsubsection{Aspect Based Sentiment Analysis}
\label{aspect}
Aspect based sentiment analysis is one of the important tasks of semantic analysis \cite{Cabasc}. In order to verify the effect of LSTA on aspect based sentiment analysis, we conducted experiments on two data sets. One was the SemEval-2014 Task 4 (SemEval14) data set\cite{SemEval14}, which contains two domains (Restaurant and Laptop). The other was the Twitter data set collected by Dong et al. \cite{Twitter}. In these two data sets, the aspect terms of each review are labeled by three sentiment polarities, which are positive, neutral and negative, respectively. For example, about an aspect term \textit{fajitas}, when it is in a sentence \textit{``I loved their fajitas, but the service is horrible."}, its polarity is positive, but for aspect term \textit{service}, its polarity is negative. Concretely, the statistics of the two data sets are provided in Table \ref{tablesem}.
\begin{table}[h]
 \caption{The statistics of the SemEval14 and Twitter data sets.}
 \smallskip
\label{tablesem}
  \centering
\setlength{\tabcolsep}{0.95mm}{
  \begin{tabular}{l|cc|cc|cc}
    \toprule
 \multirow{2}{*}{Data set} &\multicolumn{2}{c|}{Positive} &\multicolumn{2}{c}{Neutral} &\multicolumn{2}{|c}{Negative} \\
    \cmidrule(r){2-7}
     &Train & Test&  Train  & Test &Train &Test\\
 \midrule
    SemEval14(Restaurant)& 2164 & 728 & 637 & 196 &807& 196 \\
 \midrule
   SemEval14(Laptop)& 994& 341 & 464 & 169 &870& 128\\
 \midrule
   Twitter &1561 & 173 & 3127& 346 &1560& 173 \\
    \bottomrule
  \end{tabular}
}
\end{table}

In this experiment, we used Accuracy and Macro-averaged F-measure (Macro-F1) as the metrics to evaluate the effect of LSTA and the compared models~\cite{DuyuTangSentiment,Twitter}. The experimental results obtained by LSTA and the compared methods are shown in Table \ref{tablesemre}. In this table, ``Cabasc" is a content attention model for aspect based sentiment analysis \cite{Cabasc}. ``ATAE-LSTM" is an attention-based LSTM with aspect embedding, which can focus on the parts of a sentence when several aspects are taken as input \cite{ATAE-LSTM}.

From Table \ref{tablesemre}, we can see that LSTA performs best among the compared models. That is, LSTA outperforms both LSTM and previous attention models, including that apply the attention mechanism outside the cell of LSTM.\vspace{-0.3cm}

\begin{table}[h]
\caption{The experimental results obtained on the SemEval14 and Twitter data sets.}
\smallskip
\centering
\label{tablesemre}
\setlength{\tabcolsep}{0.95mm}{
 \begin{tabular}{l|cc|cc|cc}
 \toprule
 \multirow{2}{*}{Method} &\multicolumn{2}{c|}{SemEval14 (Restaurant)} &\multicolumn{2}{c}{ SemEval14 (Laptop) } &\multicolumn{2}{|c}{Twitter}   \\
 \cmidrule(r){2-7}
     &Acc.(\%) & Macro-F1&  Acc.(\%) & Macro-F1 &Acc.(\%) &Macro-F1  \\
 \midrule
   Cabasc &78.12& 0.6743 & 70.84& 0.6552 &69.51& 0.6707 \\
 \midrule
   ATAE-LSTM &77.86 & 0.6718  & 69.75& 0.6425 &69.65& 0.6762 \\
 \midrule
   LSTM &77.41 & 0.6686 & 69.74& 0.6394 &68.93& 0.6699 \\
 \midrule
   \textbf{LSTA} &$\boldsymbol{78.57}$ & $\boldsymbol{0.6801}$ & $\boldsymbol{71.16}$& $\boldsymbol{0.6559}$  &$\boldsymbol{69.94}$ & $\boldsymbol{0.6911}$ \\
 \bottomrule
 \end{tabular}
}
\end{table}\vspace{-0.5cm}

\section{Conclusion}
\label{Conclusion}

In this paper, we present a novel LSTA model to alleviate the problem that LSTM lacks the attention mechanism. The key idea behind this model is to seamlessly integrate the attention mechanism into the cell of LSTM. Experiments demonstrate that LSTA performs better than LSTM, many variants of LSTM and related attention models. Hence, LSTA can be seen as a substitute of LSTM in the sequence learning tasks.

\section*{Acknowledgment}
This work was supported by the National Key R\&D Program of China under Grant No. 2016YFC1401004, the National Natural Science Foundation of China (NSFC) under Grant No. 41706010, and No. 61876155, the Science and Technology Program of Qingdao under Grant No. 17-3-3-20-nsh, the CERNET Innovation Project under Grant No. NGII20170416, and the Fundamental Research Funds for the Central Universities of China.

\bibliographystyle{splncs04}
\bibliography{bicsbib2}
%
%
%
%
%
\end{document}